\pdfoutput=1

\documentclass[11pt]{article}

\usepackage[preprint]{acl}

\usepackage{times}
\usepackage{latexsym}
\usepackage{algorithm}
\usepackage{amssymb}
\usepackage{hyperref}
\usepackage{enumitem}
\usepackage{longtable}
\usepackage{array}
\usepackage{booktabs}
\usepackage{arydshln}
\usepackage{multirow}
\usepackage{algpseudocode}
\usepackage{amsmath}
\usepackage{subcaption}

\usepackage[T1]{fontenc}

\usepackage[utf8]{inputenc}

\usepackage{microtype}

\usepackage{inconsolata}

\usepackage{graphicx}

\title{IRIS: Interactive Research Ideation System for Accelerating Scientific Discovery}

\author{
  Aniketh Garikaparthi$^{1}$, Manasi Patwardhan$^{1}$, Lovekesh Vig$^{1}$, Arman Cohan$^{2}$ \\\\
  $^{1}$TCS Research  
  $^{2}$Yale University\\
  \texttt{\{aniketh.g}, \texttt{manasi.patwardhan}, \texttt{lovekesh.vig\}}@tcs.com,\\  \texttt{arman.cohan@yale.edu}
}

\begin{document}
\maketitle
\begin{abstract}
The rapid advancement in capabilities of large language models (LLMs) raises a pivotal question: \textit{How can LLMs accelerate scientific discovery?} This work tackles the crucial first stage of research, generating novel hypotheses.
While recent work on automated hypothesis generation focuses on multi-agent frameworks and extending test-time compute, none of the approaches effectively incorporate transparency and steerability through a synergistic Human-in-the-loop (HITL) approach. To address this gap, we introduce IRIS: Interactive Research Ideation System, an open-source platform designed for researchers to leverage LLM-assisted scientific ideation. IRIS incorporates innovative features to enhance ideation, including adaptive test-time compute expansion via Monte Carlo Tree Search (MCTS), fine-grained feedback mechanism, and query-based literature synthesis. Designed to empower researchers with greater control and insight throughout the ideation process. We additionally conduct a user study with researchers across diverse disciplines, validating the effectiveness of our system in enhancing ideation.
We  open-source
 our code \href{https://github.com/Anikethh/IRIS-Interactive-Research-Ideation-System}{here}.
\end{abstract}

\section{Introduction}

\begin{figure*}[ht]
    \centering  \includegraphics[width=\linewidth]
    {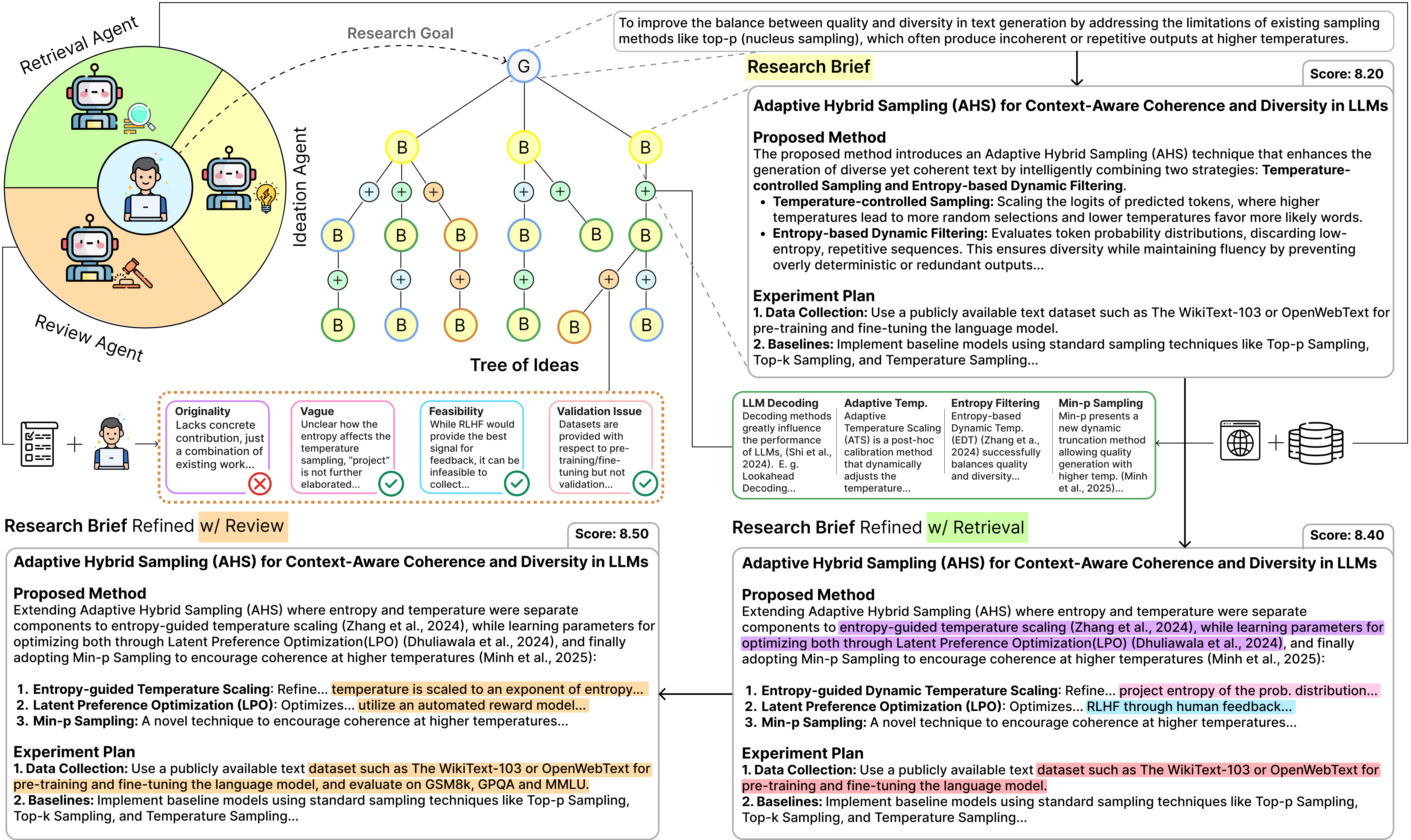}
    \caption{Human-in-the-loop Idea Generation with Monte-Carlo-Tree-Search. $\mathcal{G}$: Research Goal, $\mathcal{B}$: Research Brief}
    \label{fig:diagram}
\end{figure*}

With the growing capabilities of large language models (LLMs), the automation of scientific discovery has captured a lot of attention \cite{gridach2025agenticaiscientificdiscovery}. Agentic LLM based systems have shown potential 
of outperforming PhD researchers and postdocs on short-horizon scientific tasks like question answering, summarization and contradiction detection in various domains  \cite{skarlinski2024languageagentsachievesuperhuman, asai2024openscholarsynthesizingscientificliterature}. These advancements have spurred new opportunities of LLMs accelerating scientific discovery, which is essential given the exponential growth of scientific publications \cite{Landhuis2016ScientificLI, 10.1093/gigascience/giz053}.

Current solutions that leverage LLMs in scientific ideation primarily remain hinged on multi-agent frameworks or extending test-time compute \cite{si2024llmsgeneratenovelresearch, hu2024novaiterativeplanningsearch, ai_coscientist}, and aim to validate the quality of 
the final ideas  through human validation or LLM-as-a-judge evaluations \cite{wang-etal-2024-scimon, li2024chainideasrevolutionizingresearch, baek2025researchagentiterativeresearchidea}.
However, these approaches often fail to integrate human supervision during generation in a truly complementary manner, neglecting the nuanced expectations and goals of the user.
Consequently, despite investing significant computational resources to develop objectively ``novel'' ideas, they might not align with the user's \textit{research goals}, inevitably leading to dissatisfaction \cite{10.1145/3543758.3543761,10.1145/3640543.3645148}. 

Moreover, the importance of meaningful human intervention in the research process cannot be overstated.
Notably, AI models have been known to fabricate convincing yet 
fraudulent scientific information 
\cite{info:doi/10.2196/46924}. 
More troubling are cases of deceptive and misaligned AI behaviors \cite{anthropic2025alignment, author2025ai, betley2025emergentmisalignmentnarrowfinetuning, Baker2025CoTMonitoring}.
Recent developments of more capable Agentic LLMs have shown difficulties in transparently delegating sub-tasks, leading to \textit{"reward hacking"} behaviors \cite{anthropic2025claude}. 
In the context of idea generation, we find signs of similar \textit{"reward hacking"} where LLMs adopt fancy terminology e.g.  "Prompt Learning and Optimization Nexus" for building a library of prompts, or often proposing the use of "graphs" without any clear motivation or description behind the design choice. We observe that naive recursive feedback loops \cite{baek2025researchagentiterativeresearchidea} forcing the LLM to be more novel inevitably lead to gamifying LLM-as-a-judge metrics without adding actual value. \citet{gupta2025glittersnovelplagiarismai} carefully study the results of AI-Researcher \cite{si2024llmsgeneratenovelresearch} and advise careful assessment of LLM generated hypotheses due to signs of skillful plagiarism.
These examples highlight the pitfalls of premature reliance on fully automated systems, underscoring the need for well-designed Human-in-the-Loop (HITL) systems for scientific ideation; ensuring outcomes are accurate and aligned with human goals.



\begin{figure*}[ht]
    \centering  \includegraphics[width=\linewidth]
    {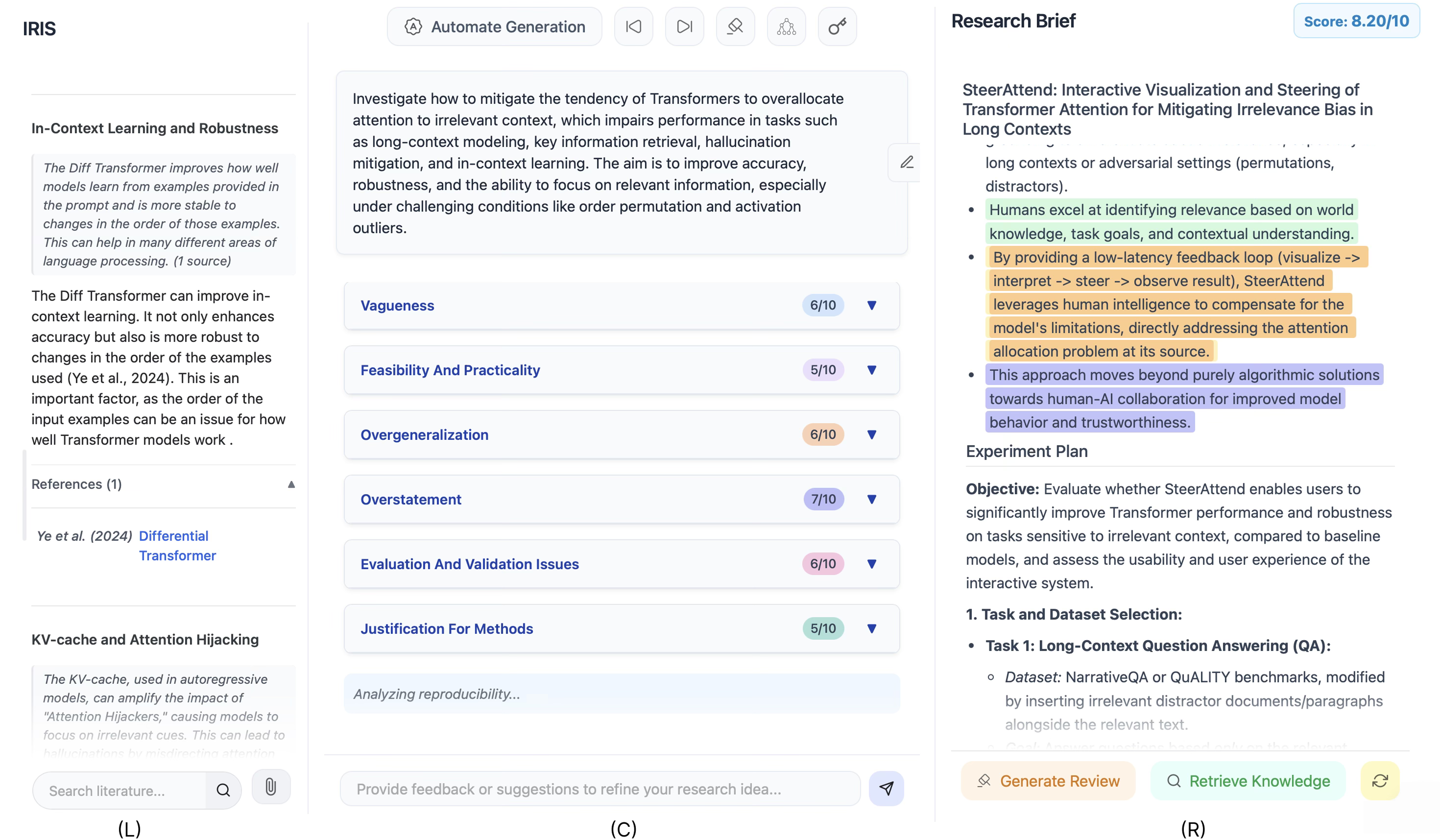}
    \caption{IRIS Platform Interface with (L) Retrieval Panel, (C) Chat Overview Panel, (R) Research Brief Panel}
    \label{fig:interface}
\end{figure*}

Despite the recent innovations made in LLM-based scientific ideation,
several key limitations persist. These include (1) generating hypotheses in a single pass \cite{si2024llmsgeneratenovelresearch}
, which overlooks the iterative nature of the ideation process. 
In contrast, \citet{pu2024ideasynthiterativeresearchidea} find that researchers typically seek to refine their hypotheses into concrete \emph{research briefs}.
(2) Optimization through feedback on coarse-grained criteria like rigorousness, originality, generalizability etc. \cite{ baek2025researchagentiterativeresearchidea}, while often critiquing entire ideas rather than specific components. (3) Simplistic retrieval augmentation such as appending keywords or abstracts of previous papers in context \cite{wang-etal-2024-scimon, si2024llmsgeneratenovelresearch},
whereas effective ideation demands a deeper, more holistic understanding of the domain literature.
(4) Unstructured and sub-optimal search of the idea space through either refinement of a generated base-idea (exploitation) \cite{wang-etal-2024-scimon, baek2025researchagentiterativeresearchidea}, or through initial search and plan (exploration) without subsequent refinement of promising ideas \cite{hu2024novaiterativeplanningsearch}. 
Finally, there is a lack of open-source implementations that would encourage broader adoption. In light of these challenges, we propose IRIS, tackling each of these limitations while enabling human intervention at every stage of the ideation process. Specifically, we make the following contributions:

\begin{itemize}
  \setlength{\itemsep}{0pt}
    \item \textbf{HITL Framework:} A user-centered design balancing human control with automation instead of entirely delegating the process of ideation to AI
    \item \textbf{Monte Carlo Tree Search:} A systematic method to iteratively explore the idea space and extend test time compute via alternating phases of exploration and exploitation (\S\ref{sec:mcts})
    \item \textbf{Fine-grained Review based Refinement:} An exhaustive taxonomy (Table \ref{tab:hierarchical_taxonomy}) with fine-grained actionable feedback for improving hypotheses (Figure \ref{fig:interface}) (\S\ref{sec:agent-arch})
    \item \textbf{Query-based Retrieval:}  Generating targeted queries for retrieving relevant literature,  with re-ranking, clustering and summarization to produce comprehensive, technical and cited responses (\S\ref{sec:agent-arch})
    \item \textbf{Open Source:} Publicly available platform for AI-Assisted scientific ideation
\end{itemize}

Finally, we conduct a user study with researchers from diverse disciplines validating the effectiveness of our designed system (\S\ref{sec:study}).

\section{Related Works} \label{sec:rel-works}

\noindent \textbf{2.1 AI Assisted Research} \label{sec:rel-works:1} \\
The integration of (AI) into scientific research has evolved from early concept-linking tools \cite{04817c26-80f1-38c3-8cf5-03c7fbde591b, 10.1145/3340531.3412684, nadkarni2021scientificlanguagemodelsbiomedical} to sophisticated systems that enhance various research stages. In recent years, LLMs have significantly transformed research life-cycles by assisting in literature searches \cite{zheng2024openresearcherunleashingaiaccelerated, ajith2024litsearchretrievalbenchmarkscientific, asai2024openscholarsynthesizingscientificliterature}, citation recommendations \cite{9917887, ZHANG2022115826,  press2024citemelanguagemodelsaccurately}, review of scientific documents \cite{zhou-etal-2024-llm}, experimental design \cite{huang2024mlagentbenchevaluatinglanguageagents, schmidgall2025agentlaboratoryusingllm}, scientific claim verification \cite{lu2024aiscientistfullyautomated}, theorem proving \cite{song2025leancopilotlargelanguage}, manuscript writing \cite{weng2025cycleresearcherimprovingautomatedresearch}, and reading assistants\footnote{\href{https://jenni.ai/}{JenniAI}, \href{https://typeset.io/}{SciSpace}, \href{https://scholarai.io/}{ScholarAI}}.

\noindent \textbf{2.2 Human-AI Co-creation Systems}\\
The emergence of Gen AI has introduced a new dimension to Co-creation systems, setting them apart from previous ones where machines primarily served as supportive tools for human users \cite{Davis2015, muller2020mixed, 10.1145/3613904.3642466}. Recent studies, such as those by \citet{kar80484, 10.1145/3613904.3642698}, demonstrate the effectiveness of Gen AI tools in creative tasks, particularly through their steerability and explainability. This has led to growing emphasis among researchers to develop design guidelines for integrating Gen AI into existing frameworks \cite{10.1145/3290605.3300233, 10.1145/3419764}. 
We build IRIS for researcher-in-the-loop ideation while incorporating design principles from prior work, such as minimizing opacity, adopting granular feedback, encouraging AI processing delays \cite{10.1145/3290605.3300233, 10.1145/3613904.3642698}, and replacing rigid post-hoc analysis with oversight across planning, generation, and retrospection stages \cite{10.1145/3419764}.



\noindent \textbf{2.3 Automated Hypothesis Generation}\\
\citet{10.1145/2623330.2623667} demonstrate the first proof of principle for automated hypothesis generation through text mining of scientific literature, leveraging techniques such as entity detection and graph-based diffusion of information. Rising capabilities of text completion models has driven significant advancements in this field \cite{wang-etal-2024-scimon, lu2024aiscientistfullyautomated, li2024chainideasrevolutionizingresearch, hu2024novaiterativeplanningsearch, si2024llmsgeneratenovelresearch, kumar2024largelanguagemodelsunlock, baek2025researchagentiterativeresearchidea, ai_coscientist}. However, current efforts focus on fully automated systems, often overlooking the critical role of human involvement.
\emph{Acceleron} demonstrates one of the first human-in-the-loop (HITL) framework assisting researchers in validation of motivation behind a research problem and synthesizing a method for the same  \cite{nigam-etal-2024-interactive}, followed by \citet{pu2024ideasynthiterativeresearchidea} making an  attempt to develop an interactive idea generation system.
These approaches remain limited, allowing idea exploration only within a predefined framework, restricting flexibility and adaptability. Furthermore, their system lacks sophisticated components like automated fine-grained feedback,  literature retrieval targeted to the research goal and scaling test-time compute. 

\section{IRIS}

Broadly, the system expects as input a research goal $\mathcal{G}$ consisting of a research problem and it's motivation, and outputs a research brief $\mathcal{B}$ consisting of a Title, Proposed Methodology and Experiment Plan, while improving it's quality; either in \emph{semi-automatic} manner through directions from the researcher or \emph{autonomously} exploiting Monte Carlo Tree Search (MCTS). We provide detailed overview of our system including the implementation of agents (\S\ref{sec:agent-arch}) and MCTS adaptation for hypothesis generation (\S\ref{sec:mcts}).

\subsection{Agent Architecture}\label{sec:agent-arch}

IRIS employs a three-agent architecture consisting of an ideation agent, a review agent, and a retrieval agent. 
The ideation agent navigates the search space of possible research ideas, while the review and retrieval agents provide feedback and relevant scientific context respectively.\\ 

\textbf{Ideation Agent} generates and iteratively improves the research brief. It can toggle between a \emph{semi-automatic} mode, to receive guidance from a researcher to refine research briefs through steering reviews, retrievals or employing custom feedback, and  a completely \emph{autonomous} mode to explore and exploit the idea space by leveraging \emph{actions} which support iterative refinement of the research briefs through MCTS.

\textbf{Review Agent} is accountable for two tasks namely providing \textit{reward} and \textit{feedback}. For evaluation of an idea, we have defined a hierarchical taxonomy of aspects grounded in real-world scientific critique (For example, \cite{10.1371/journal.pone.0259238}, \cite{kennard-etal-2022-disapere}, \cite{dycke-etal-2023-nlpeer}), detailed in Table \ref{tab:hierarchical_taxonomy}. Review Agent is auto-triggered after each new generation of the research brief to provide a \textit{reward} averaged over the scores assigned to distinct aspects, based on the evaluation provided for the complete research brief. 

As opposed to the parallel works \cite{wang-etal-2024-scimon, baek2025researchagentiterativeresearchidea} that focus on coarse-level criteria and provide broad evaluation of the entire generated research brief, usually, a feedback with respect to an aspect is  applicable to only specific parts of the research brief. For example, only some component of the brief can be infeasible or some other component requires more clarity.
Addressing this need, when explicitly triggered by the researcher, the review agent switches to a fine-grained evaluation, delivering targeted, actionable feedback
on each aspect of the taxonomy for distinct segments of the current research-brief 
(Figures  \ref{fig:diagram} and \ref{fig:interface} (R) ).
This fine-grained feedback is verified by the researcher and omitted if deemed irrelevant. Then the review agent computes reward  based on the scores of the verified aspects of the feedback.  This adept human intervention coupled with granular feedback, successfully mitigates \textit{``reward hacking''} behavior of LLMs.





\textbf{Retrieval Agent: }
For the input research goal, the retrieval agent synthesizes queries targeted to retrieve literature relevant to the research goal. For answering each query, 
it adopts Ai2 Scholar QA API\footnote{\url{https://allenai.org/blog/ai2-scholarqa}}.  The pipeline consists of two-stage retrieval followed by three-stage generation. The Semantic Scholar API's (storing over 200M open access papers) snippet search endpoint \cite{kinney2023semanticscholaropendata}  extracts relevant passages, which are re-ranked to retain top-k passages and aggregated at the paper level. With the finalized set of passages, the retrieval agent (i) extracts quotes from the passages relevant to the query, (ii) generates a plan to produce an organized report with sections, and clusters the top-k passages accordingly, and (iii) generates cited sections-wise reports along with summaries (Figure \ref{fig:interface} (L)).
Our motivation for adopting ScholarQA stems from the limitations of naive RAG failing to appropriately answer global questions targeted at a corpus  as opposed to a single document \cite{edge2025localglobalgraphrag}. We also provide the ability for the researcher to upload papers in the form of PDF documents, which they think to be relevant but have been missed out as the part of the retrieval.  The retrieval agent parses the PDF through Grobid
based doc2json tool\footnote{\url{https://github.com/allenai/s2orc-doc2json}} and appends the most relevant chunks to the  context for the ideation agent to refine the research brief. 



\subsection{Monte Carlo Tree Search Framework} \label{sec:mcts}
To systematically explore the vast space of potential research ideas,
IRIS employs Monte Carlo Tree Search (MCTS) \cite{10.1007/11871842_29}. MCTS allows the system to effectively extend test-time compute similar to recent work in augmenting LLM reasoning \cite{qi2024mutualreasoningmakessmaller, guan2025rstarmathsmallllmsmaster}. Unlike applications with objective rewards (e.g., mathematics, code generation), scientific ideation quality is subjective. We adapt MCTS by using the LLM-based Review Agent as a proxy judge to estimate the \textit{quality} (reward) of generated hypotheses.

Formally,  given a research goal $\mathcal{G}$, our system constructs a search tree $\mathcal{T}$ rooted with $\mathcal{G}$.
A state $s$ encapsulates the current \{research brief $b$, reward estimate $r$, latest review feedback $f$ (if applicable, else $\phi$), and retrieved knowledge $k$ (if applicable, else $\phi$)\}. Edges represent actions $a$ taken by the Ideation Agent to transition between states.
We define a 
comprehensive action space $\mathcal{A}$ = \{$a_1$: generate, $a_2$: refine w/ retrieval, $a_3$: refine w/ review, $a_4$: refine w/ user feedback\}. 
The MCTS process iteratively builds the tree over $N$ iterations, guided by the Upper Confidence Bound for Trees (UCT) algorithm \cite{coquelin2007banditalgorithmstreesearch}.
UCT of a node $n$ is defined by: 
\begin{equation} \label{eq:ucb_revised}
\text{UCT}(n) = \frac{Q(n)}{N(n)} + c \sqrt{\frac{\ln N(n_p)}{N(n)}}
\end{equation}
where $Q(n)$ is the total reward  at child node $n$ accumulated from its children, $N(n)$ is its visit count, $N(n_p)$ is the visit count of the parent node of $n$ , and $c$ is the exploration constant. Algorithm \ref{alg:mcts_standard_flow} outlines the MCTS process. Each node $n$ stores its state $s_n$ as defined above,  $Q(n)$ and $N(n)$.

\begin{algorithm}[h]
\caption{MCTS for Research Idea Generation}
\label{alg:mcts_standard_flow}
\begin{algorithmic}[1]
\Require Research goal $\mathcal{G}$, iterations $N$, max depth $d_{\text{max}}$, actions $\mathcal{A}$, constant $c$
\State Initialize tree $\mathcal{T}$ with root $n_0$ (state $s_0=\mathcal{G}$, $Q(n_0)=0$, $N(n_0)=0$).
\For{$i = 1$ to $N$}
    \State $n_{\text{leaf}} \gets$ \Call{Select}{$n_0, c$}
    \State $r \gets$ \Call{Evaluate}{$n_{\text{leaf}}$} 
    \If{depth < $d_{\text{max}}$}
        \State \Call{Expand}{$n_{\text{leaf}}, \mathcal{A}$} 
    \EndIf
    \State \Call{Backpropagate}{$n_{\text{leaf}}, r$} 
\EndFor
\State \Return \Call{BestChild}{$n_0$} 
\end{algorithmic}
\end{algorithm}

Each iteration involves four phases:

\textbf{\textsc{Select}($n_{root}, c$)}: 
Traverse the tree from the root $n_{0}$ to select a leaf node $n_{\text{leaf}}$. At each node $n$ during traversal, if $n$ has any unvisited children ($Q(n) = 0$), one such child is randomly selected.
If all children of $n$ have been visited, the next node is chosen by:
$\arg\max_{n' \in \text{children}(n)} (\text{UCT}(n'))$.

\textbf{\textsc{Evaluate}($n_{\text{leaf}}$)}: Obtain reward $r$ for the state $s_{\text{leaf}}$ of $n_{\text{leaf}}$ via the Review Agent.

\textbf{\textsc{Expand}($n_{\text{leaf}}, \mathcal{A}$)}: If $n_{\text{leaf}}$ is non-terminal and below $d_{\text{max}}$, create child nodes $n'$ for each applicable action $a \in \mathcal{A}$, with $Q(n')=0, N(n')=0$.

\textbf{\textsc{Backpropagate}($n_{\text{leaf}}, r$)}: Update $Q$ and $N$ values for $n_{\text{leaf}}$ and its ancestors with reward $r$.

\textbf{\textsc{BestChild}($n_0$)}: After $N$ iterations, select the child of $n_0$ with the highest average reward $Q/N$.\\

\noindent \textbf{Memory:} 
Agents maintain trajectory-level memory. For instance, the Ideation Agent recalls 
generated briefs,
the Retrieval Agent remembers past queries, and the Review Agent tracks prior feedback. This helps steer the generation towards non-redundant refinements.

\noindent \textbf{Cost:} MCTS can be computationally intensive. IRIS incorporates budget controls, allowing users to set limits. For tighter budgets, the system prioritizes exploitation by lowering the exploration constant $c$, ensuring delivery of few refined outputs rather than numerous low-quality ones.

\section{Evaluation} \label{sec:study}
To assess the effectiveness and usability of IRIS, we conduct automated evaluations and user studies. 

\subsection{Experiment Setup}

\textbf{System Implementation:} IRIS's user interface is developed using HTML, CSS, JavaScript. The core LLM functionalities are powered by Gemini-2.0-Flash \cite{google2024gemini} accessed via LiteLLM\footnote{\url{https://docs.litellm.ai/docs/}}, which allows users to substitute other LLMs of their choice. We utilize Gemini's built-in safety filters to mitigate harmful or inappropriate queries.

\textbf{Metrics:} We employ LLM-as-a-judge, popularly adopted in parallel literature \cite{baek2025researchagentiterativeresearchidea, ai_coscientist}. We use two methods guided by our pre-defined criteria (Table \ref{tab:hierarchical_taxonomy}). \textit{absolute score:} each generated hypothesis (1-10), and
\textit{relative score:} aggregating head-to-head comparisons and preferences to compute ELO ratings. 

To contextualize the alignment of LLM-as-a-judge with human preferences in the context of scientific ideation, we prompt baselines Gemini-2.0-Flash, ChatGPT, ChatGPT w/ search and Claude 3.5 Haiku to generate novel research briefs. Then ask users and LLMs to rate the generations in the order of their preference.

\subsection{User Study} \label{sec:usr-study}

We conducted a user study with 8 researchers (N=8) from diverse fields (AI/NLP, Chem, Physics, HCI) and experience levels. Two users voluntarily participated twice (10 total case studies). Each $\sim$60 min session involved: 1) Defining a research goal, 2) Blindly ranking initial set of hypotheses, 3) Interacting with IRIS, 4) Completing a post-task survey.

\subsection{Results and Analysis}

\textbf{Metric Validation:} Human baseline rankings correlated moderately with LLM based ELO scores (Pearson's r=0.60) but weakly with LLM based absolute scores (r=0.45). With this learning we plan to replace the LLM-as-the-judge scores, displayed to showcase the quality of the idea, with the ELO ratings.

\begin{figure}[htbp]
    \centering 
    \begin{subfigure}[b]{0.48\linewidth}
        \centering \includegraphics[width=\linewidth]{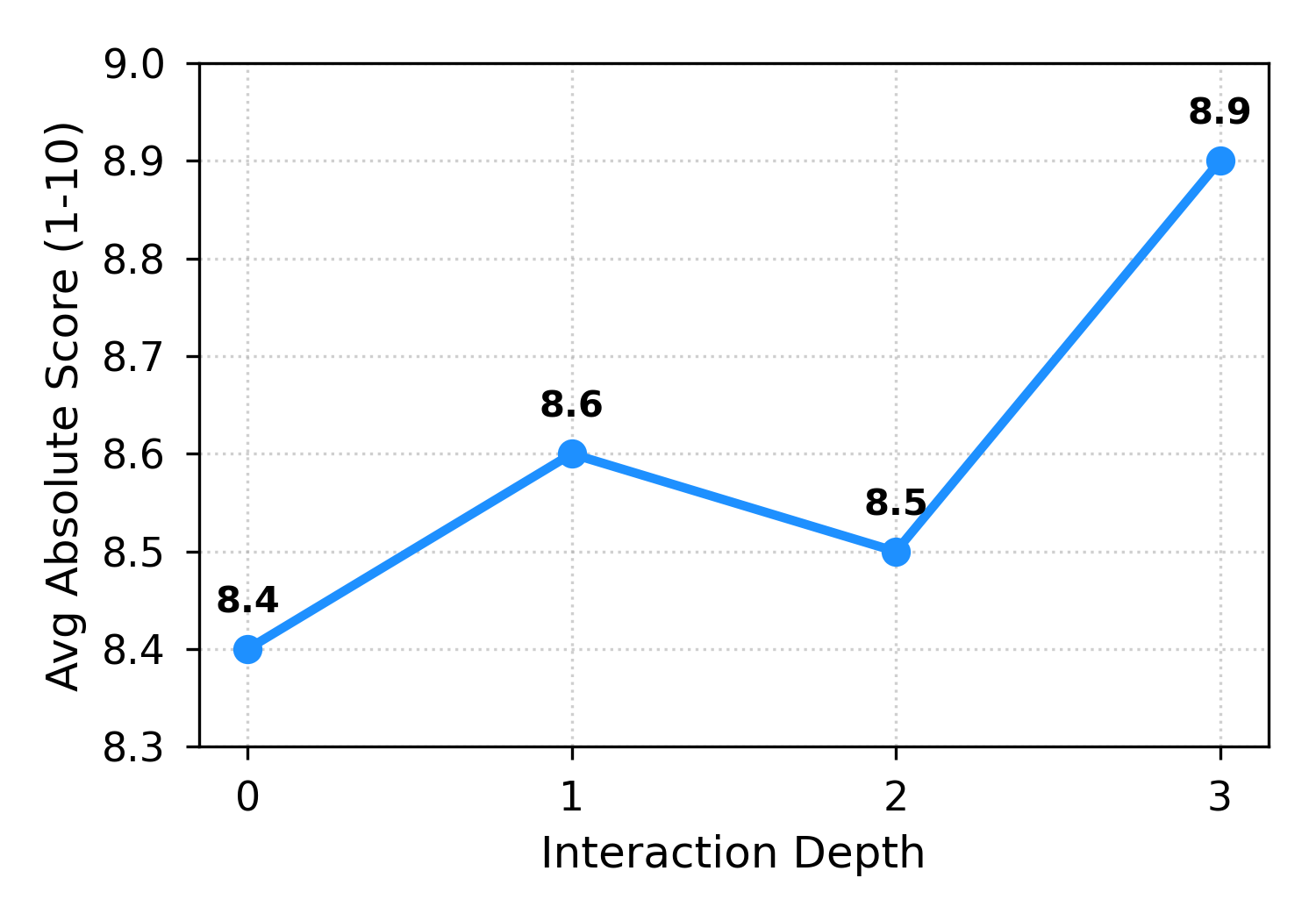}
        \caption{Absolute Score Improvement.}
        \label{fig:iterative-abs}
    \end{subfigure}
    \hfill
    \begin{subfigure}[b]{0.48\linewidth}
        \centering \includegraphics[width=\linewidth]{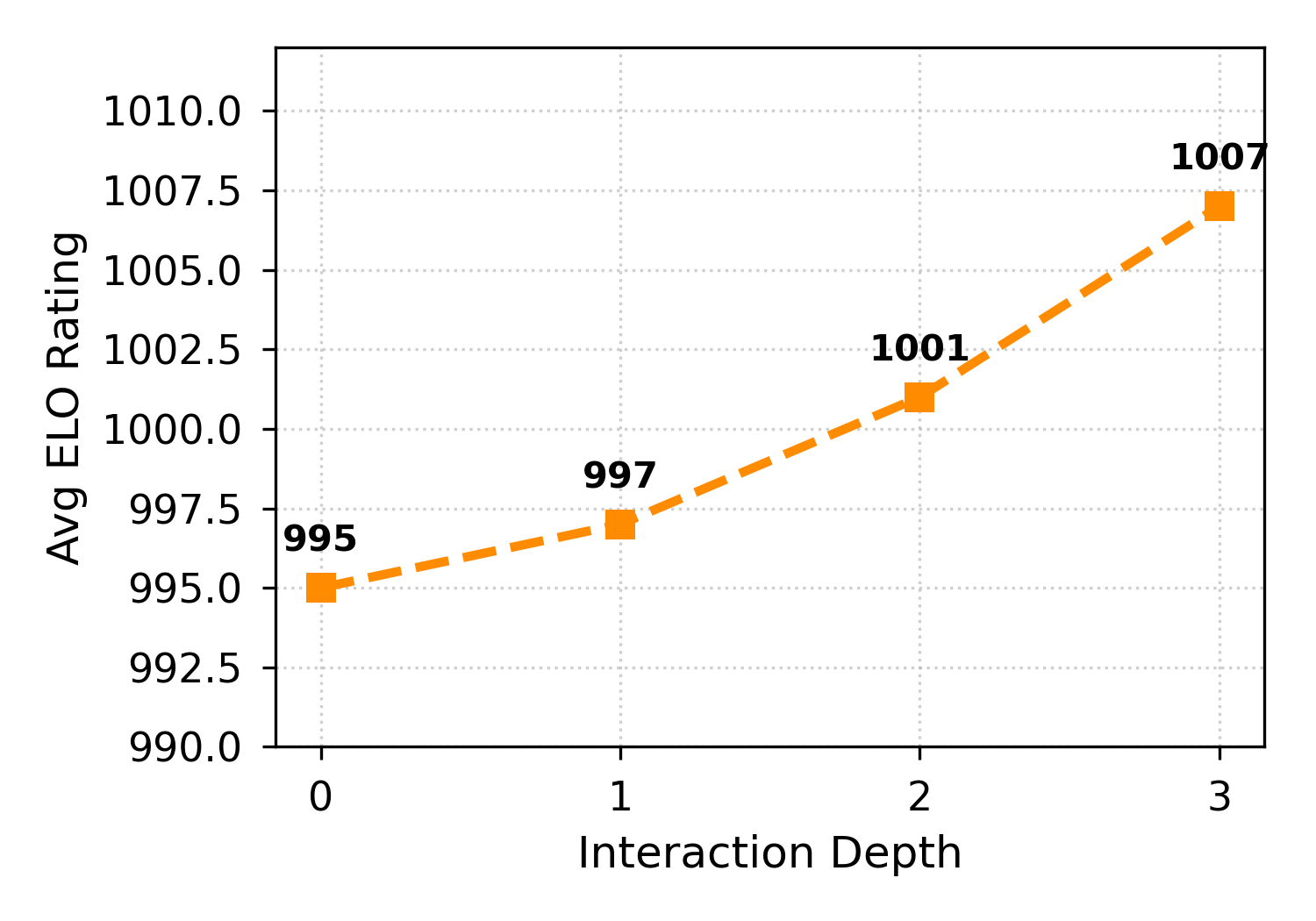}
        \caption{ELO Rating Improvement.}
        \label{fig:iterative-elo} 
    \end{subfigure}
    
    \caption{Iterative improvement in hypothesis quality within IRIS over interaction depth (up to depth 3). Interaction enhances both absolute scores and ELO ratings.}
    \label{fig:iterative-improvements}
\end{figure}

\textbf{Automated Evaluation:} 
LLM-as-a-judge evaluations (Figure~\ref{fig:iterative-improvements}) showed that user interaction within IRIS consistently improved hypothesis quality, increasing average absolute scores by 0.5 points and ELO ratings by 12 points for a tree depth of 3.

\textbf{User Study Feedback:} Quantitative ratings (Table~\ref{tab:user-ratings}) show users found the fine-grained feedback highly insightful and unpromptedly mentioned better usability and control over other reading assistant interfaces mentioned in \S\ref{sec:rel-works}.

\begin{table}[ht]
\centering
\vspace{-2mm} 
\resizebox{\linewidth}{!}{
\begin{tabular}{lc}
\toprule
\textbf{Feature / Aspect} & \textbf{Mean Rating (± Std Dev)} \\
\midrule
Usefulness of Fine-grained Feedback & 4.3 ± 0.7 \\
MCTS Tree Interface (Steerability) & 4.2 ± 0.6 \\
Quality of Lit. Summaries & 3.7 ± 0.8 \\
Usability and control & 4.5 ± 0.7 \\
Overall Satisfaction (Final Research Brief) & 3.9 ± 0.7 \\
\bottomrule
\end{tabular}%
}
\caption{User ratings (1-5 Likert scale) for key IRIS features and overall satisfaction (N=10).}
\label{tab:user-ratings}
\vspace{-3mm} 
\end{table}

Additionally, through qualitative feedback we arrived at the following insights: 
\begin{itemize}
    \setlength{\itemsep}{0pt}
    \item \textbf{Steerability:} All users valued the MCTS tree for control and transparency over ideation.
    \item \textbf{Feedback:} Critiques often reflected user's own concerns (87.5\% users) and sometimes sparked novel insights (50\% cases).
    \item \textbf{Retrieval:} Found to be facilitating grounding of ideas, but quality varied with domains such as chemistry and physics research, matching the lower rating (3.7/5).
    We attribute this to reduced availability of relevant literature in the semantic scholar corpus.
    \item \textbf{Relevance:} hypotheses often shared similarities with or extended users' ongoing work (62.5\% users).
\end{itemize}

\textbf{Overall Improvement:} Post-interaction, 25\% (2/8) found the hypothesis substantially better, 50\% (4/8) marginally better, and 25\% (2/8) similar quality. Crucially, all users reported enhanced understanding of the proposed methodology, and considered it to be promising.

\section{Conclusion} \label{sec:conclusion}
We introduce IRIS, an Interactive Research Ideation System, to augment automated scientific hypothesis generation with human expertise. We apply MCTS to iteratively explore the idea space, refine ideas with fine-grained segment level reviews and targeted query based multi-document retrieval; offering a steerable environment for researchers during LLM-driven scientific ideation.
Our user study validates the usability and effectiveness of our system, demonstrating consistent improvement in hypothesis quality increasing average absolute scores by 0.5 points and ELO ratings by 12
points for a tree depth of 3. Crucially, users frequently considered the generated hypotheses plausible and worthy of further investigation. We position that the potential of LLMs, particularly within human-AI collaborative frameworks, 
for developing novel scientific hypothesis remains a heavily underexplored avenue. We present IRIS as a concrete step towards realizing this untapped potential.

\section*{Limitations} \label{sec:lim}
Currently the system relies on the researcher as the judge to verify the quality of the emerging idea at each iteration, augmented by LLM-as-the-judge.  This reliance is based on the assumption of sufficient domain expertise of the researcher. As opposed to this in future we aim for a true Human AI Co-creation System, where more foundational LLMs with scientific expertise, questions researchers for the choices he or she has made leading to a two way socratic review and refinement communication, simulating a more realistic scenario of brain-storming between  colleagues or a mentor and a mentee. 

Due to budget constraints, we have not explored frontier LLMs such as Claude 3.7 Sonnet, Grok-3 or reasoning models like Gemini-2.5-Pro,  o1 etc. The quality of produced hypothesis in terms of novelty and effectiveness would likely benefit from stronger base models.


\bibliography{latex/acl_latex}

\appendix

\onecolumn

\section{Review Taxonomy}

\label{sec:appendix}

\begin{longtable}{>{\centering\arraybackslash}m{3cm} m{5cm} p{7cm}}

\toprule

\textbf{Aspect} & \textbf{Sub-aspect} & \textbf{Definition} \\

\midrule

\textbf{Originality} & Lack of Novelty & The idea does not introduce a significant or meaningful advancement over existing work, lacking originality or innovation. \\
               \cdashline{2-3}[5pt/5pt]
               & Assumptions & The idea relies on untested or unrealistic assumptions that may weaken its validity or applicability. \\
\midrule

\textbf{Clarity} & Vagueness & The idea is presented in an unclear or ambiguous manner, making it difficult to understand its core components or contributions. \\
                 \cdashline{2-3}[5pt/5pt]
                 & Contradictory Statements & The idea contains internal inconsistencies or conflicts in its assumptions, methods, or conclusions. \\
                 \cdashline{2-3}[5pt/5pt]
                 & Alignment & The idea is not aligned with the problem statement and its objectives. \\
\midrule

\textbf{Feasibility} & Feasibility and Practicality & The idea is not practical or achievable given current technological, theoretical, or resource constraints. \\
                      \cdashline{2-3}[5pt/5pt]
                      & Justification for Methods & The idea does not provide sufficient reasoning or evidence to explain why specific methods, techniques, or approaches were chosen. \\
\midrule

\textbf{Effectiveness} & Evaluation and Validation Issues & The idea lacks rigorous evaluation methods, such as insufficient benchmarks, inadequate baselines, or poorly defined success metrics. \\
                        \cdashline{2-3}[5pt/5pt]
                        & Reproducibility and Robustness & The idea does not provide sufficient detail or transparency to allow others to replicate or verify its findings, and is not resilient to variations in input data, assumptions, or environmental conditions. The degree to which the solution consistently produces accurate and dependable results is low, making it less reliable. \\
\midrule

\textbf{Impact} & Overgeneralization and Overstatement & The idea extends its conclusions or applicability beyond the scope of the context provided or exaggerates its claims, significance, or potential impact beyond what is supported by evidence or reasoning. \\
                \cdashline{2-3}[5pt/5pt]
                & Impact & The idea is not impactful or significant. It does not solve a real problem. It does not create value by solving a significant problem or fulfilling a need for individuals, organizations, or society. \\
                \cdashline{2-3}[5pt/5pt]
                & Ethical and Social Considerations & The idea does not adhere to ethical standards and is harmful to individuals, communities, or the environment. \\

\bottomrule

\caption{Hierarchical Review Taxonomy}

\label{tab:hierarchical_taxonomy}

\end{longtable}

\begin{figure*}[htbp] 
    \centering 

    \begin{subfigure}[b]{0.45\linewidth} 
        \centering
        \includegraphics[width=\linewidth]{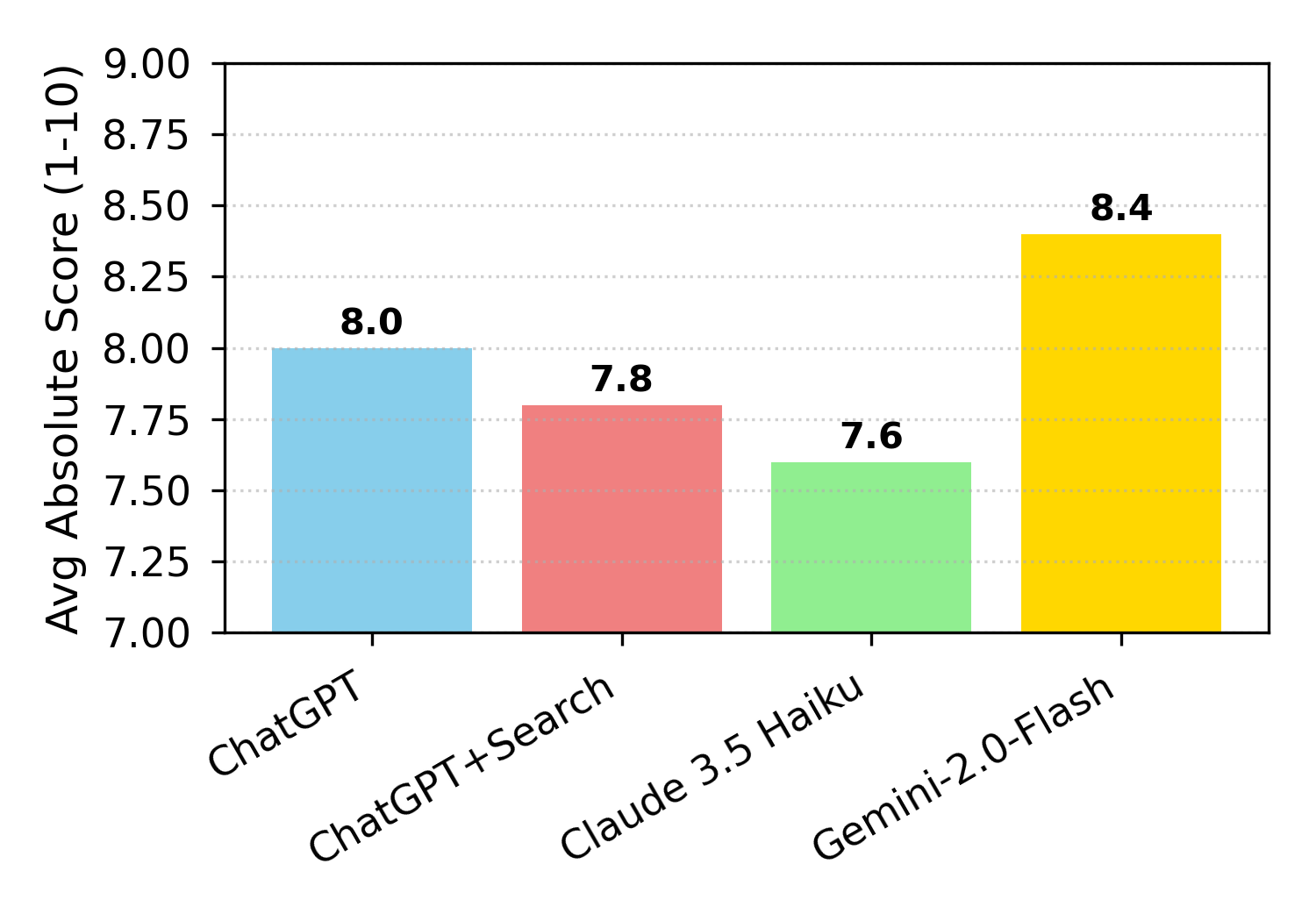}
        \caption{Baseline Comparison (Absolute Score).}
        \label{fig:abs-score-comp}
    \end{subfigure}
    \hfill 
    \begin{subfigure}[b]{0.45\linewidth} 
        \centering
        \includegraphics[width=\linewidth]{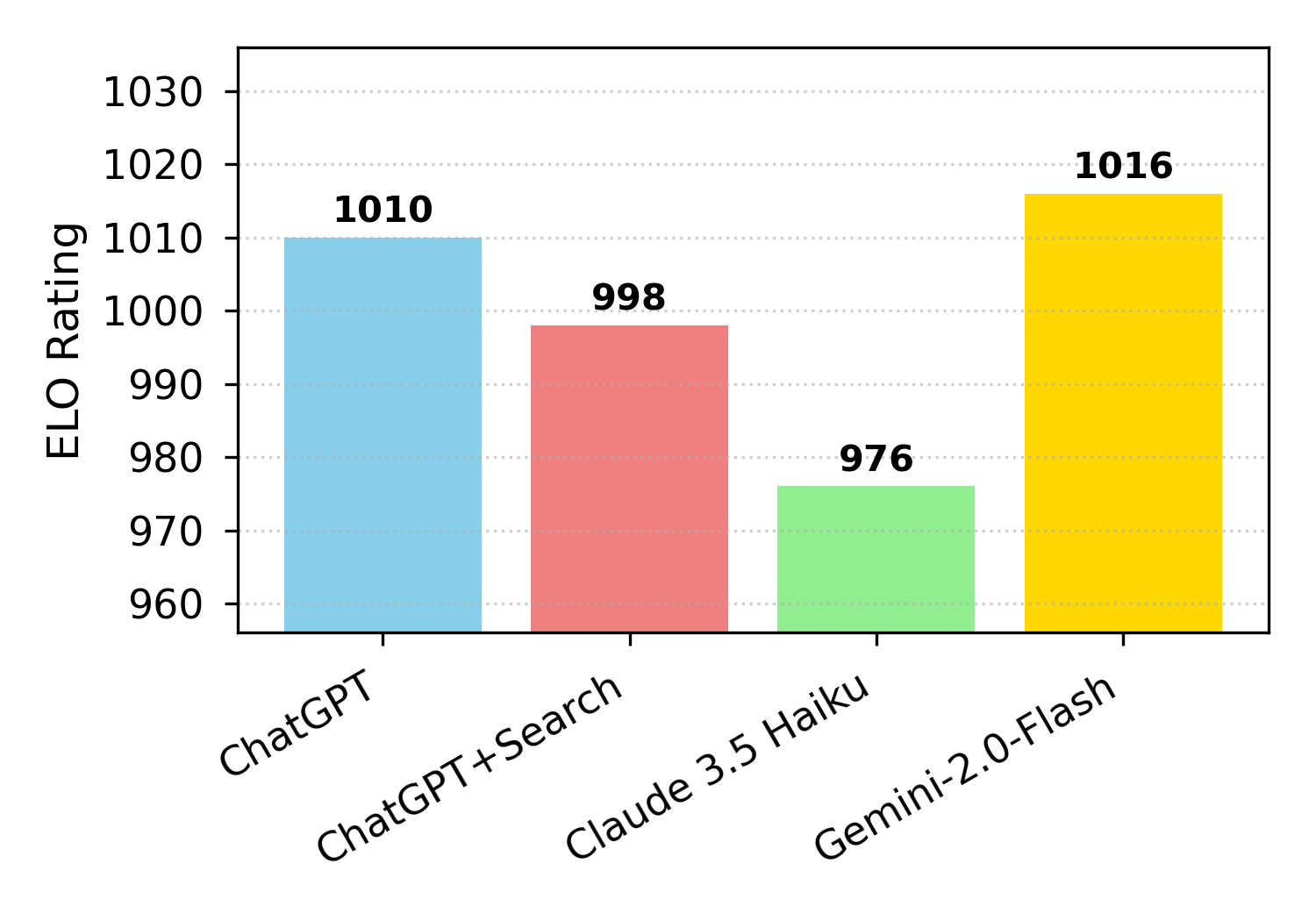}
        \caption{Baseline Comparison (ELO Rating).}
        \label{fig:elo-score-comp}
    \end{subfigure}

    \vspace{75pt} 

    \includegraphics[width=0.9\linewidth]{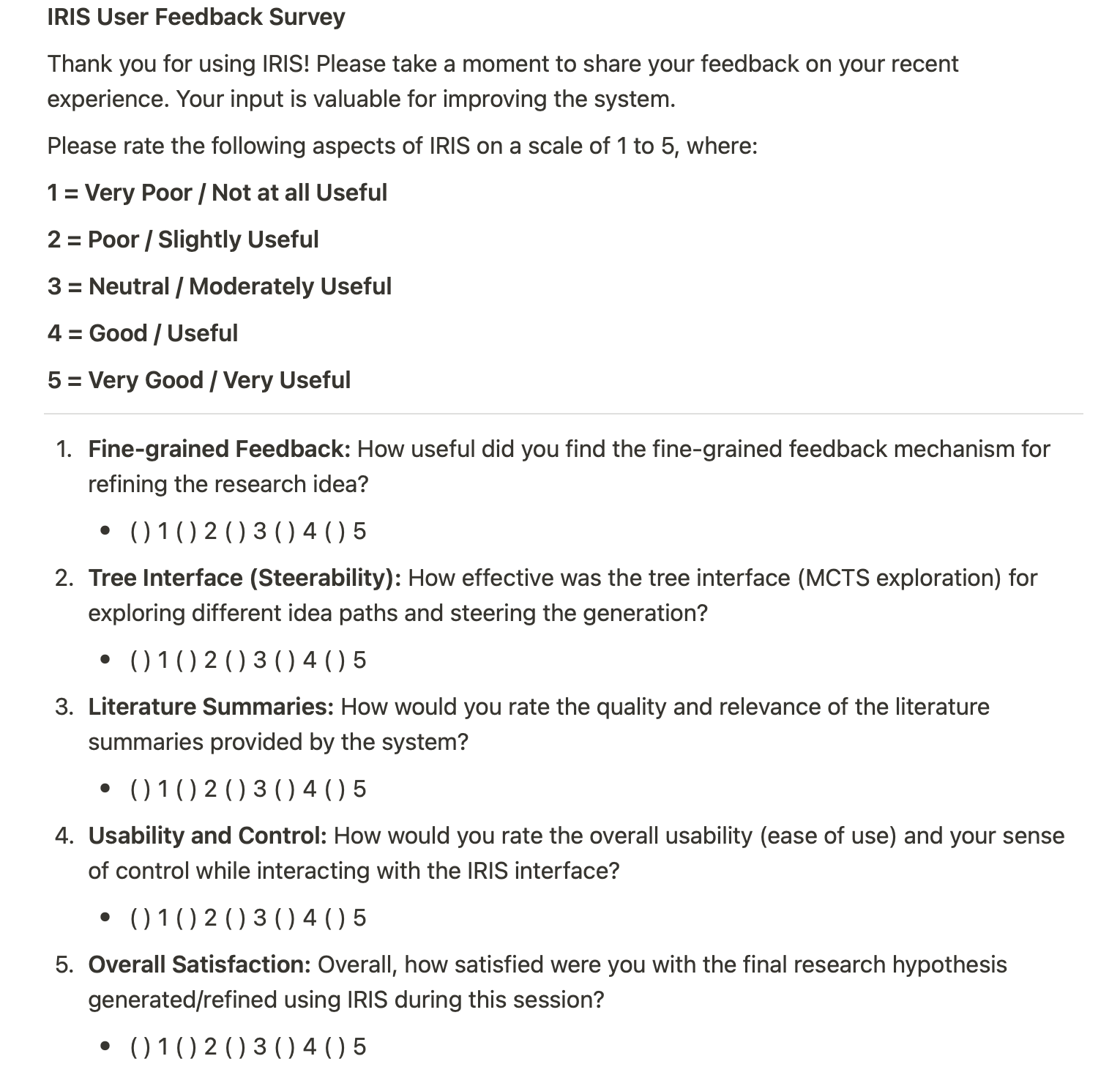} 

    \caption{Top: Comparison of hypothesis quality generated by baseline methods (ChatGPT, ChatGPT+Search, Claude 3.5 Haiku, Gemini-2.0-Flash) using LLM-as-a-judge absolute scores and ELO ratings. Bottom: User Survey Feedback Form Questions.}
    \label{fig:combined-appendix-plots} 
\end{figure*}

\end{document}